\documentclass[runningheads]{llncs}

\usepackage{cite}
\usepackage[colorlinks=true,
    linkcolor=blue,
    urlcolor=blue,citecolor=blue]{hyperref}
\usepackage{amsmath}
\usepackage{multirow}
\usepackage{algorithmic}
\usepackage{array}
\usepackage{amssymb}
\usepackage{booktabs}
\usepackage{stfloats}
\usepackage{url}
\usepackage{xcolor}
\usepackage{xspace}
\usepackage{bm}
\usepackage{pifont} 

\usepackage[misc]{ifsym}
\usepackage[T1]{fontenc}

\usepackage{graphicx}

\begin{document}

\title{TheraAgent: Multi-Agent Framework with Self-Evolving Memory and Evidence-Calibrated Reasoning for PET Theranostics}

\titlerunning{TheraAgent for PET/CT Theranostics}

%

  

\author{Zhihao Chen\inst{1} \and
Jiahui Wang\inst{2} \and
Yizhou Chen\inst{2} \and
Xiaozhong Ji\inst{3} \and
Xiaobin Hu\inst{4} \and
Jimin Hong\inst{2} \and
Wolfram Andreas Bosbach\inst{2} \and
Axel Rominger\inst{2} \and
Ali Afshar-Oromieh\inst{2} \and
Hongming Shan\inst{1}\textsuperscript{(\Letter)} \and
Kuangyu Shi\inst{2}\textsuperscript{(\Letter)}
}
\authorrunning{Z. Chen et al.}

\institute{Fudan University \and
University of Bern \and
Nanjing University \and
National University of Singapore}

\maketitle

\newcommand{\etal}{\textit{et al}.\xspace}
\newcommand{\ie}{\textit{i}.\textit{e}.\xspace}
\newcommand{\eg}{\textit{e}.\textit{g}.\xspace}
\newcommand{\tabincell}[1]{\begin{tabular}[l]{@{}l@{}} #1\end{tabular}}

\newcommand{\mat}[1]{\boldsymbol{#1}}
\newcommand{\vct}[1]{\boldsymbol{#1}}

\newcommand{\std}[1]{\tiny #1}

\newcommand{\modelname}{TheraAgent\xspace}
\newcommand{\amem}{SEA-Mem\xspace}

\begin{abstract}
PET theranostics is transforming precision oncology, yet treatment response varies substantially—many patients receiving $^{177}$Lu-PSMA radioligand therapy (RLT) for metastatic castration-resistant prostate cancer (mCRPC) fail to respond, demanding reliable pre-therapy prediction.
While LLM-based agents have shown remarkable potential in complex medical diagnosis, their application to PET theranostic outcome prediction remains unexplored, which faces three key challenges: 
(1) \textit{data and knowledge scarcity}: RLT was only FDA-approved in 2022, yielding few training cases and insufficient domain knowledge in general LLMs; 
(2) \textit{heterogeneous information integration}: robust prediction hinges on structured knowledge extraction from PET/CT, laboratory tests, and free-text clinical documentation; (3) \textit{evidence-grounded reasoning}: clinical decisions must be anchored in trial evidence rather than LLM hallucinations. 
In this paper, we present \modelname, to our knowledge, the first agentic framework for PET theranostics, with three core innovations: 
(1) \textbf{Multi-Expert Feature Extraction with Confidence-Weighted Consensus} where three specialized experts process heterogeneous inputs with uncertainty quantification; (2) \textbf{Self-Evolving Agentic Memory (\amem)} that learns prognostic patterns from accumulated cases, enabling case-based reasoning from limited data; (3) \textbf{Evidence-Calibrated Reasoning} integrating a curated theranostics knowledge base to ground predictions in VISION/TheraP trial evidence. Evaluated on 35 real patients and 400 synthetic cases, \modelname achieves 75.7\% overall accuracy on real patients and 87.0\% on synthetic cases, outperforming MDAgents and MedAgent-Pro by over 20\%.
These results highlight a promising blueprint for trustworthy AI agents in PET theranostics, enabling trial-calibrated, multi-source decision support. Code will be released upon acceptance.
\keywords{PET Theranostics \and Multi-Agent \and Self-Evolving Memory \and Evidence-Calibrated Reasoning.}
\end{abstract}

\section{Introduction}

PET theranostics integrates diagnostic imaging with targeted therapy, representing a transformative approach in precision oncology~\cite{mehrens2023cost}. 
$^{177}$Lu-PSMA radioligand therapy (RLT), which received FDA approval in 2022 for patients with PSMA positive metastatic castration-resistant prostate cancer (mCRPC), uses PSMA PET/CT for patient screening as they share the same targeting mechanism for imaging and therapeutic delivery~\cite{sartor2021lutetium,mehrens2023cost}. 
The VISION trial demonstrated significant overall survival benefits, yet only 46\% of patients achieved PSA response~\cite{sartor2021lutetium}, while TheraP reported 66\% PSA response in mCRPC patients with progression following docetaxel treatment~\cite{hofman2021therap}. This variability implies that many patients may undergo multiple costly cycles with cumulative radiation exposure yet achieve limited benefit, underscoring the need for reliable pre-therapy prediction of treatment response to guide individualized decision-making~\cite{yadav2019radioligand}.

Recent advances in large language model (LLM) based agents have shown remarkable potential in complex medical diagnosis~\cite{hu2025landscape, wang2026deepmed, shi2026medxiaohe}. 
MDAgents~\cite{kim2024mdagents} introduces adaptive multi-agent collaboration for medical decision-making, while MedAgent-Pro~\cite{wang2025medagent} proposes hierarchical agentic workflows for evidence-based diagnosis. 
Memory-augmented approaches~\cite{gao2025survey} such as MemGPT~\cite{packer2023memgpt,xu2025amem} enable long-term knowledge retention. 
However, these frameworks focus on general medical reasoning and have not been applied to theranostics, where outcome prediction faces three fundamental challenges: 
(1) Data and Knowledge Scarcity: RLT is a nascent therapy with limited documented cases and scarce curated knowledge; general-purpose LLMs lack sufficient theranostic domain expertise, making robust learning and reasoning difficult.
(2) Heterogeneous Information Integration. Reliable prediction requires structured extraction and fusion of PSMA PET/CT imaging, laboratory values, and free-text clinical documentation, often spanning multiple languages and formats.
(3) Evidence-Grounded Reasoning. Clinical decisions must be anchored in quantitative trial evidence (e.g., VISION and TheraP) rather than hallucinated explanations, yet most agents lack mechanisms to calibrate predictions against such evidence.

\begin{figure*}[t]
\centering
\includegraphics[width=1\textwidth]{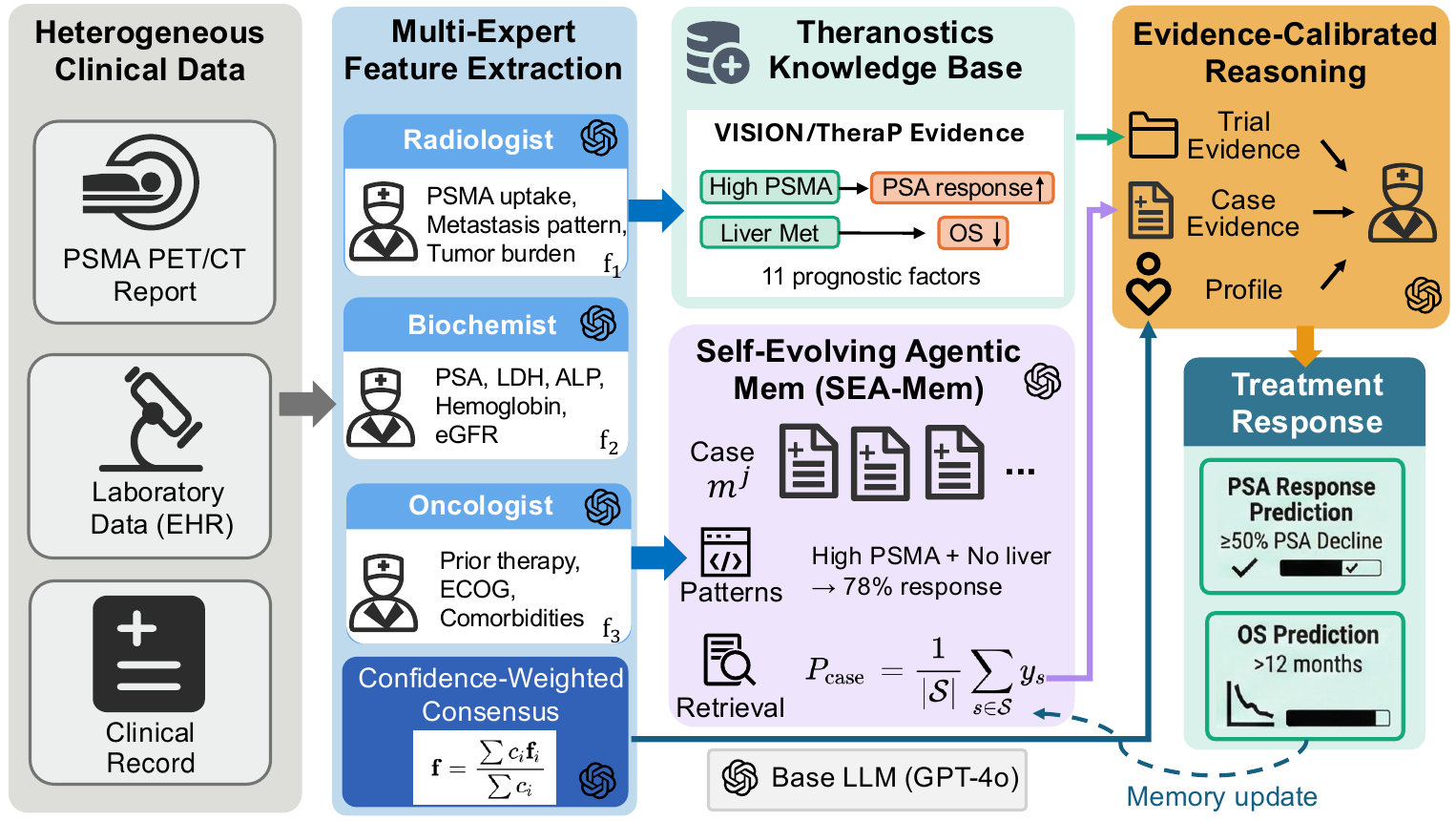}
\caption{Overview architecture of \modelname}
\label{fig:framework}
\end{figure*}

In this paper, we present \modelname, to our knowledge the first LLM-based agentic framework for PET theranostics (Figure~\ref{fig:framework}), which addresses above challenges with targeted innovations: (1) \textbf{Multi-Expert Feature Extraction}: Three specialized experts---Radiologist, Biochemist, and Oncologist---process heterogeneous multilingual inputs in parallel, each optimized for its data modality (imaging reports, laboratory values, clinical narratives), with confidence-weighted consensus for robust feature fusion.
(2) \textbf{Self-Evolving Agentic Memory (\amem)}: Unlike static knowledge bases, \amem learns prognostic patterns from accumulated cases and updates when treatment outcomes become available. This enables case-based reasoning from limited data while continuously adapting to institutional experience.
(3) \textbf{Evidence-Calibrated Reasoning}: We construct a domain-specific knowledge base of 23 curated publications on Lu-177 PSMA therapy under guidance from nuclear medicine physicians. The final reasoning module retrieves relevant trial evidence from this knowledge base alongside similar cases from \amem, synthesizing all sources to generate predictions with explicit citations.

Evaluated on 35 real patients and 400 synthetic cases, \modelname achieves accuracy of 75.7\% on real data and 87.0\% on synthetic data, significantly outperforming single LLM and state-of-the-art medical agents, with especially strong gains on misleading cases where surface features contradict true prognostic factors.
Beyond methodological contributions, our work establishes a promising paradigm for trustworthy AI-assisted clinical decision support in PET theranostics, potentially reducing unnecessary treatment cycles and improving patient outcomes through more accurate pre-therapy stratification.

\section{Methodology}

\subsection{Problem Formulation}
For each patient, we obtained three types of pre-treatment documentation: (1) PSMA PET/CT reports containing radiotracer distribution, SUV measurements, and lesion descriptions across bone, lymph node, and visceral sites; (2) laboratory examinations including PSA level, renal function (creatinine, eGFR), liver enzymes (ALP, LDH), and hematological parameters (hemoglobin, platelets); 
(3) clinical records documenting diagnosis history, prior treatment lines (ADT, chemotherapy), ECOG performance status, and comorbidities. 

Given these heterogeneous inputs, we predict: (1) PSA response, defined as $\geq$50\% PSA decline within 2 treatment cycles, and (2) overall survival $>$12 months post-treatment. Unlike static prediction models, \modelname can \textit{evolve} its knowledge when actual outcomes $y^*$ become available, enabling continuous improvement from clinical feedback.

\subsection{Multi-Expert Feature Extraction with Confidence-Weighted Consensus}
Reliable outcome prediction requires extracting structured information from three distinct data sources: imaging reports, laboratory values, and clinical narratives. Each source has its own format, terminology, and potential ambiguities. We address this heterogeneity by deploying three specialized LLM-based expert agents that process their respective modalities in parallel:

\begin{itemize}
    \item \textbf{Radiologist Agent} parses PSMA PET/CT reports to identify PSMA expression level (high/moderate/low/heterogeneous), metastasis distribution across bone, lymph node, and visceral sites, and qualitative tumor burden.
    \item \textbf{Biochemist Agent} extracts laboratory values---PSA and its trend, hemoglobin, alkaline phosphatase (ALP), lactate dehydrogenase (LDH), and renal function (eGFR)---from structured and semi-structured EHR data.
    \item \textbf{Oncologist Agent} interprets clinical narratives to identify prior treatment lines (ADT, chemotherapy), ECOG performance status, and relevant comorbidities, handling multilingual documentation when necessary.
\end{itemize}

Each expert outputs a structured feature profile $\mathbf{f}_i$ (containing extracted clinical attributes in JSON format) along with a confidence score $c_i \in [0,1]$ reflecting extraction certainty.

\noindent\textbf{Confidence-Weighted Consensus.}\quad
A fourth agent, the \textbf{Integrator}, receives all expert outputs and synthesizes them into a unified patient representation. The Integrator is prompted to weight each expert's contribution according to their confidence scores---giving more consideration to high-confidence extractions while treating uncertain findings with appropriate caution. When experts disagree, the Integrator resolves conflicts by examining the underlying evidence. The output is a unified patient profile $\mathbf{f}$ that serves as input to both memory retrieval and reasoning stages.

\subsection{Self-Evolving Agentic Memory (\amem)}
With only 35 documented cases in our dataset, traditional supervised learning is infeasible. 
To address this data scarcity, we introduce \amem, a lightweight memory module inspired by recent work on agentic memory systems~\cite{park2023generative,xu2025amem}. The key idea is to accumulate processed cases over time and learn prognostic patterns that can inform predictions for new patients.

\noindent\textbf{Memory Structure.}\quad
Each entry  $m^{(j)} = (\mathbf{f}^{(j)}, y^{(j)})$ stores the extracted feature profile and documented outcome (PSA response, OS $>$12 months) for a historical patient. Entries are indexed by key prognostic features---PSMA level, presence of liver/lung metastases, and prior chemotherapy---to enable efficient retrieval.

\noindent\textbf{Pattern Learning.}\quad
When three or more cases share a common feature combination, \amem computes the empirical outcome rate for that subgroup. For instance, among 8 patients with high PSMA and no liver metastasis, 7 achieved PSA response, yielding a learned pattern: ``high PSMA + no liver metastasis $\rightarrow$ 77.8\% PSA response.'' These patterns are stored alongside raw cases and retrieved when relevant.

\noindent\textbf{Case-Based Retrieval.}\quad
Given a new patient with profile $\mathbf{f}_{\text{query}}$, \amem retrieves the top-$k$ most similar historical cases by weighted feature matching. The retrieval returns: (1) the similar cases with their outcomes, providing concrete reference points; and (2) any matching learned patterns. The empirical outcome rate is computed as:
$ P_{\text{case}} = \frac{1}{|\mathcal{S}|}\sum_{s \in \mathcal{S}} y_s $, where $\mathcal{S} = \text{top-}k\text{ similar cases}$.

\noindent\textbf{Memory Update.}\quad
When treatment outcomes become available, new cases are added to memory, and existing patterns are updated with the fresh data. This enables \modelname to improve over time as the institution accumulates more experience with RLT.

\subsection{Evidence-Calibrated Reasoning}
Clinical predictions must be grounded in established evidence, not LLM hallucinations. 
We integrate two evidence sources: trial evidence from curated clinical literature and case evidence from institutional experience (\amem).

\noindent\textbf{Trial Evidence: Knowledge Base Construction.}\quad
Under the guidance of nuclear medicine physicians, we constructed a knowledge base comprising 23 peer-reviewed publications on Lu-177 PSMA therapy. The corpus centers on landmark trials (VISION~\cite{sartor2021lutetium}, TheraP~\cite{hofman2021therap}), systematic reviews, and outcome studies. From these sources, we extracted 11 key prognostic factors with documented associations---for example, high PSMA expression correlates with improved PSA response, while liver metastasis predicts worse survival. Each document is chunked and indexed for semantic retrieval.

\noindent\textbf{Trial Evidence Retrieval.}\quad
Given the patient's profile $\mathbf{f}$, we construct targeted queries based on extracted clinical characteristics (\eg, ``PSMA expression and PSA response,'' ``liver metastasis prognosis''). The retrieval system returns relevant passages such as: ``TheraP: high PSMA uptake achieved 66\% PSA response vs 37\% in low-uptake group''; ``VISION: liver metastasis HR=2.1 for overall survival.''

\noindent\textbf{Reasoning Module.}\quad
The final prediction synthesizes three information sources, shown in Figure~\ref{fig:framework}: (1) the patient's profile $\mathbf{f}$; (2) case evidence from \amem---similar patients and learned patterns; and (3) trial evidence from the knowledge base. 
The LLM integrates all sources and cites them explicitly. 
For example: ``This patient has high PSMA expression and bone-only metastases [\textit{Profile}]. 
Among 5 similar cases, 4 achieved PSA response [\textit{Case Evidence}]. 
TheraP shows high PSMA correlates with 66\% response; absence of visceral metastases is favorable, VISION HR=0.67 [\textit{Trial Evidence}]. Prediction: positive PSA response.'' 
This design ensures predictions are traceable to specific sources.

\noindent\textbf{Pipeline Summary.}\quad
The complete workflow: (1) three expert agents extract features, unified by the Integrator; (2) parallel retrieval of case evidence from \amem and trial evidence from the knowledge base; (3) the reasoning module synthesizes profile, case evidence, and trial evidence to generate predictions with explicit citations.

\begin{table}[t]
\caption{Performance comparison on real clinical data (n=35 patients). \modelname significantly outperforms all baselines.}
\label{tab:main_results}
\centering
\resizebox*{0.98\linewidth}{!}{
\begin{tabular*}{1\linewidth}{@{\extracolsep{\fill}}lcccccc}
\toprule[1.5pt]
\multirow{2}{*}{\textbf{Method}} & \multicolumn{2}{c}{\textbf{PSA Response}} & \multicolumn{2}{c}{\textbf{OS $>$12m}} & \multicolumn{2}{c}{\textbf{Overall}} \\
\cmidrule(lr){2-3} \cmidrule(lr){4-5} \cmidrule(lr){6-7}
& Acc (\%) & F1 (\%) & Acc (\%) & F1 (\%) & Acc (\%) & F1 (\%) \\
\midrule
\multicolumn{7}{l}{\textit{Single LLM}} \\
GPT-4o & 48.6 & 30.8 & 48.6 & 43.8 & 48.6 & 37.3 \\
\quad + CoT & 57.1 & 68.1 & 48.6 & 40.0 & 52.9 & 54.0 \\
\quad + Self-Consistency & 40.0 & 36.4 & 54.3 & 42.9 & 47.1 & 39.6 \\
\quad + RAG & 65.7 & 66.7 & 65.7 & 70.0 & 65.7 & 68.3 \\
\midrule
\multicolumn{7}{l}{\textit{Medical AI Agents}} \\
MedAgents & 48.6 & 62.5 & 60.0 & 72.0 & 54.3 & 67.3 \\
MDAgents & 60.0 & 66.7 & 65.7 & 70.6 & 62.9 & 68.6 \\
MedAgent-Pro & 54.3 & 58.8 & 60.0 & 66.7 & 57.1 & 62.8 \\
\midrule
\textbf{\modelname (Ours)} & \textbf{77.1} & \textbf{81.2} & \textbf{74.3} & \textbf{82.9} & \textbf{75.7} & \textbf{82.1} \\
\bottomrule[1.5pt]
\end{tabular*}
}
\end{table}

\section{Experiments}

\subsection{Dataset and Experimental Setup}
\noindent\textbf{Real Clinical Data.}\quad
We collected data from 35 mCRPC patients who underwent multiple cycles of $^{177}$Lu-PSMA-617 RLT at * hospital.
Each sample includes PSMA PET/CT reports, medical records, and laboratory examinations.
For PSA response prediction, 18 patients  achieved $\geq$50\% PSA decline; for OS prediction, 22 patients survived $>$12 months.

\noindent\textbf{Synthetic Data.}\quad
Given the scarcity of clinical data, we generate 400 synthetic patients to enable more comprehensive evaluation. 
Following evidence-based synthesis practices~\cite{chen2021synthetic}, our pipeline consists of: (1) patient profiles sampled from VISION/TheraP trial statistics (PSA distributions, metastasis rates, outcome probabilities); (2) clinical documents generated by Claude Opus 4.6 thinking using domain-specific templates, following LLM-based simulation practices~\cite{li2024agent,park2023generative}; (3) multi-expert verification for consistency. 
Cases are stratified by complexity: clear (50\%), ambiguous (30\%), and misleading (20\%), where 168 patients achieved $\geq$50\% PSA decline and 224 patients survived $>$12 months.

\noindent\textbf{Baselines.}\quad
We compare against single LLM methods (direct prompting, Chain-of-Thought~\cite{wei2022chain}, Self-Consistency~\cite{wang2022self}, RAG) and state-of-the-art medical agents (MDAgents~\cite{kim2024mdagents}, MedAgent-Pro~\cite{wang2025medagent}, MedAgents~\cite{tang2024medagents}).

\noindent\textbf{Implementation Details.}\quad
For fair comparison, all methods use GPT-4o~\cite{achiam2023gpt} as the base LLM. For RAG baselines, we use the same theranostics knowledge base as \modelname. Our framework is implemented using LangChain for the RAG components. 

\begin{table}[t]
\centering
\caption{Accuracy (\%) on synthetic data by case complexity.}
\label{tab:synthetic_results}
\resizebox*{1\linewidth}{!}{
\begin{tabular*}{1\linewidth}{@{\extracolsep{\fill}}lcccc}
\toprule[1.5pt]
\textbf{Method} & \textbf{Clear} & \textbf{Ambiguous} & \textbf{Misleading} & \textbf{Overall} \\
& (n=200) & (n=120) & (n=80) & \\
\midrule
GPT-4o & 79.5 & 63.3 & 40.0 & 66.8 \\
GPT-4o w. RAG & 89.5 & 70.0 & 55.0 & 76.8 \\
MDAgents & 86.0 & 66.7 & 51.3 & 73.3 \\
MedAgent-Pro & 83.0 & 68.3 & 47.5 & 71.5 \\
\midrule
\textbf{\modelname} & \textbf{95.0} & \textbf{79.2} & \textbf{78.8} & \textbf{87.0} \\
\bottomrule[1.5pt]
\end{tabular*}
}
\end{table}

\subsection{Main Results}

Table~\ref{tab:main_results} shows that \modelname achieves 75.7\% overall accuracy on real clinical data, outperforming MDAgents (62.9\%) and MedAgent-Pro (57.1\%) by over 13\%. Existing agents lack theranostics-specific knowledge, while \modelname combines case evidence from \amem with trial evidence for accurate prognostic factor identification.
Table~\ref{tab:synthetic_results} reveals significant improvement on misleading cases (78.8\% vs.\ 40.0--55.0\%), where \modelname correctly identifies favorable prognosis by grounding reasoning in trial evidence rather than surface features.

\begin{table}[h]
\centering
\caption{Ablation study on real data. All values are accuracy (\%).}
\label{tab:ablation}
\begin{tabular*}{0.88\linewidth}{@{\extracolsep{\fill}}lrrr}
\toprule[1.5pt]
\textbf{Configuration} & \textbf{PSA} & \textbf{OS} & \textbf{Overall} \\
\midrule
\modelname (full) & \textbf{77.1} & \textbf{74.3} & \textbf{75.7} \\
\quad w/o Multi-Expert & 62.9 & 68.6 & 65.7 \\
\quad w/o \amem & 68.6 & 71.4 & 70.0 \\
\quad w/o Evidence-Calibrated & 71.4 & 74.3 & 72.9 \\
GPT-4o (Single LLM) & 48.6 & 48.6 & 48.6 \\
\bottomrule[1.5pt]
\end{tabular*}
\end{table}

\subsection{Ablation Study}

Table~\ref{tab:ablation} validates the effectiveness of each component: Multi-Expert contributes 10.0\%, \amem contributes 5.7\%, and Evidence-Calibrated Reasoning contributes 2.8\%. The multi-expert design shows the largest gain, confirming that specialized agents effectively parse heterogeneous clinical data (PET reports, lab values, clinical narratives). Removing \amem causes notable degradation, demonstrating that case evidence effectively compensates for the scarcity of theranostics data. The full system achieves 27.1\% improvement over GPT-4o, validating that all three components contribute effectively.

\subsection{Case Study: Interpretable Reasoning}

Figure~\ref{fig:case_study} illustrates how \modelname's evidence-calibrated reasoning leads to correct predictions where baselines fail. For Patient G1\_06, a 68-year-old mCRPC patient with high PSMA uptake and bone-only metastases, both GPT-4o and MedAgents incorrectly predicted negative outcomes. In contrast, \modelname correctly identified favorable prognostic factors by grounding its reasoning in trial evidence: (1) high PSMA expression correlates with 66\% PSA response rate; (2) absence of visceral metastases is associated with better OS (HR=0.48). The \underline{underlined evidence} demonstrates how evidence-calibrated reasoning enables interpretable and accurate predictions.

\begin{figure}[t]
\centering
\fbox{\parbox{0.95\linewidth}{
\footnotesize
\textbf{Patient G1\_06:} 68y male, mCRPC, ECOG 1, PSA 45.2, SUVmax 32.5, bone-only mets\\
\textbf{Ground Truth:} PSA Response: \textcolor{green!60!black}{\textbf{Yes}} \quad OS$>$12m: \textcolor{green!60!black}{\textbf{Yes}}\\[3pt]
\hrule
\vspace{3pt}
\textbf{GPT-4o:} ``Bone metastases indicate advanced disease.'' $\rightarrow$ PSA: \textcolor{red}{\ding{55}} OS: \textcolor{red}{\ding{55}}\\[3pt]
\textbf{MedAgents:}\\
\textit{Agent 1:} ``Multiple bone mets suggest high burden.'' \textit{Agent 2:} ``Prior chemo failure is negative.''\\
\textit{Consensus:} ``Poor prognosis expected.'' $\rightarrow$ PSA: \textcolor{red}{\ding{55}} OS: \textcolor{red}{\ding{55}}\\[3pt]
\hrule
\vspace{3pt}
\textbf{\modelname:}\\
\textit{Profile:} High PSMA, bone-only mets, ECOG 1, prior ADT+docetaxel\\
\textit{Trial Evidence:} \underline{High PSMA $\rightarrow$ better response; No liver $\rightarrow$ favorable OS}\\
\textit{Case Evidence:} 4/5 similar cases achieved PSA response\\
$\rightarrow$ Prediction: PSA: \textcolor{green!60!black}{\ding{51}} \quad OS: \textcolor{green!60!black}{\ding{51}} \quad \textbf{(Correct)}
}}
\caption{Case study. MedAgents discusses but lacks theranostics knowledge; \modelname grounds predictions in trial evidence and case evidence.}
\label{fig:case_study}
\end{figure}

\section{Conclusion}

We present \modelname, the first multi-agent LLM framework for PET theranostic outcome prediction. By integrating multi-expert feature extraction, self-evolving agentic memory, and evidence-calibrated reasoning, our framework demonstrates the feasibility of applying LLM-based agents to this clinically important yet underexplored domain. Experimental results show that \modelname outperforms existing medical agents, particularly on cases where surface features contradict true prognostic factors, validating the necessity of evidence grounding for reliable prediction. Ultimately, accurate pre-therapy stratification could reduce unnecessary treatment cycles and improve patient outcomes by guiding individualized decision-making.

We acknowledge that current performance leaves room for improvement due to task complexity and data scarcity; future work will focus on multi-center datasets, reinforcement learning from clinical feedback~\cite{xia2025mmedagent}, and prospective validation.


\begin{thebibliography}{10}
\providecommand{\url}[1]{\texttt{#1}}
\providecommand{\urlprefix}{URL }
\providecommand{\doi}[1]{https://doi.org/#1}

\bibitem{achiam2023gpt}
Achiam, J., Adler, S., Agarwal, S., Ahmad, L., Akkaya, I., Aleman, F.L., Almeida, D., Altenschmidt, J., Altman, S., Anadkat, S., et~al.: Gpt-4 technical report. arXiv preprint arXiv:2303.08774  (2023)

\bibitem{chen2021synthetic}
Chen, R.J., Lu, M.Y., Chen, T.Y., Williamson, D.F.K., Mahmood, F.: Synthetic data in machine learning for medicine and healthcare. Nature Biomedical Engineering  \textbf{5},  493--497 (2021). \doi{10.1038/s41551-021-00751-8}

\bibitem{gao2025survey}
Gao, H.a., Geng, J., Hua, W., et~al.: A survey of self-evolving agents: On path to artificial super intelligence. arXiv preprint arXiv:2507.21046  \textbf{1} (2025)

\bibitem{hofman2021therap}
Hofman, M.S., Emmett, L., Sandhu, S., Iravani, A., Joshua, A.M., Goh, J.C., Pattison, D.A., et~al.: {[}$^{177}${Lu]{]}Lu-PSMA-617 versus cabazitaxel in patients with metastatic castration-resistant prostate cancer ({TheraP}): a randomised, open-label, phase 2 trial}. Lancet Oncology  \textbf{22}(8),  1175--1186 (2021)

\bibitem{hu2025landscape}
Hu, X., Qian, Y., Yu, J., Liu, J., Tang, P., Ji, X., Xu, C., Liu, J., Yan, X., Yu, X., et~al.: The landscape of medical agents: A survey. Authorea Preprints  (2025)

\bibitem{kim2024mdagents}
Kim, Y., Kim, D., Kim, J., Kang, J., Lee, J.: {MDAgents}: An adaptive collaboration of {LLMs} for medical decision-making. In: Advances in Neural Information Processing Systems ({NeurIPS}) (2024), preprint: arXiv:2404.15155

\bibitem{li2024agent}
Li, J., Li, X., Zhang, Y., et~al.: Agent hospital: A simulacrum of hospital with evolvable medical agents. arXiv preprint  (2024)

\bibitem{mehrens2023cost}
Mehrens, D., Kramer, K.K., Unterrainer, L.M., et~al.: Cost-effectiveness analysis of 177lu-psma-617 radioligand therapy in metastatic castration-resistant prostate cancer. Journal of the National Comprehensive Cancer Network  \textbf{21}(1),  43--50 (2023)

\bibitem{packer2023memgpt}
Packer, C., Fang, V., Patil, S.G., Gonzalez, J.E., Holzman, A.: {MemGPT}: Towards {LLMs} as operating systems. arXiv preprint  (2023)

\bibitem{park2023generative}
Park, J.S., O'Brien, J.C., Cai, C.J., Morris, M.R., Liang, P., Bernstein, M.S.: Generative agents: Interactive simulacra of human behavior. In: Proceedings of the ACM Symposium on User Interface Software and Technology (UIST) (2023). \doi{10.1145/3586183.3606763}, preprint: arXiv:2304.03442

\bibitem{sartor2021lutetium}
Sartor, O., de~Bono, J., Chi, K.N., Fizazi, K., Herrmann, K., Rahbar, K., Tagawa, S.T., et~al.: Lutetium-177--{PSMA}-617 for metastatic castration-resistant prostate cancer. The New England Journal of Medicine  \textbf{385}(12),  1091--1103 (2021). \doi{10.1056/NEJMoa2107322}

\bibitem{shi2026medxiaohe}
Shi, B., Cui, B., Jiang, B., et~al.: Medxiaohe: A comprehensive recipe for building medical mllms. arXiv preprint arXiv:2602.12705  (2026)

\bibitem{tang2024medagents}
Tang, X., Wang, Y., Zhao, Y., et~al.: {MedAgents}: Large language models as collaborators for zero-shot medical reasoning. In: Findings of the Association for Computational Linguistics (2024), preprint: arXiv:2311.10537

\bibitem{wang2026deepmed}
Wang, H., Feng, S., Yang, X., et~al.: Deepmed: Building a medical deepresearch agent via multi-hop med-search data and turn-controlled agentic training \& inference. arXiv preprint arXiv:2601.18496  (2026)

\bibitem{wang2022self}
Wang, X., Wei, J., Schuurmans, D., Le, Q., Chi, E., Narang, S., Chowdhery, A., Zhou, D.: Self-consistency improves chain of thought reasoning in language models. In: International Conference on Learning Representations (ICLR) (2023), preprint: arXiv:2203.11171

\bibitem{wang2025medagent}
Wang, Z., Xie, Y., Chen, X., et~al.: {MedAgent-Pro}: Evidence-based multi-modal medical diagnosis. In: ICLR (2026)

\bibitem{wei2022chain}
Wei, J., Wang, X., Schuurmans, D., Bosma, M., et~al.: Chain-of-thought prompting elicits reasoning in large language models. In: Advances in Neural Information Processing Systems ({NeurIPS}) (2022), preprint: arXiv:2201.11903

\bibitem{xia2025mmedagent}
Xia, P., Wang, J., Peng, Y., Zeng, K., Wu, X., Tang, X., Zhu, H., Li, Y., Liu, S., Lu, Y., et~al.: {MMedAgent-RL}: Optimizing multi-agent collaboration for multimodal medical reasoning. arXiv preprint arXiv:2506.00555  (2025)

\bibitem{xu2025amem}
Xu, W., et~al.: {A-MEM}: Agentic memory for {LLM} agents. arXiv preprint  (2025)

\bibitem{yadav2019radioligand}
Yadav, M.P., Ballal, S., Sahoo, R.K., Dwivedi, S.N., Bal, C.: Radioligand therapy with 177lu-psma for metastatic castration-resistant prostate cancer: a systematic review and meta-analysis. American Journal of Roentgenology  \textbf{213}(2),  275--285 (2019)

\end{thebibliography}

\end{document}